\documentclass[runningheads]{llncs}
\usepackage{comment}
\usepackage{array}
\usepackage{graphicx}

\usepackage{hyperref}

\usepackage[T2A]{fontenc}
\usepackage[utf8]{inputenc}
\usepackage[russian,english]{babel}

\begin{document}
\title{Advances of Transformer-Based Models for News Headline Generation}
\titlerunning{Transformers for Headline Generation}
%
\author{Alexey Bukhtiyarov, Ilya Gusev}
\authorrunning{}
%
\institute{Moscow Institute of Physics and Technology, Moscow, Russia\\
\email{\{bukhtiyarov.ao,ilya.gusev\}@phystech.edu}}
\maketitle              
\begin{abstract}
\selectlanguage{english} 
Pretrained language models based on Transformer architecture are the reason for recent breakthroughs in many areas of NLP, including sentiment analysis, question answering, named entity recognition. Headline generation is a special kind of text summarization task. Models need to have strong natural language understanding that goes beyond the meaning of individual words and sentences and an ability to distinguish essential information to succeed in it. In this paper, we fine-tune two pretrained Transformer-based models (mBART and BertSumAbs) for that task and achieve new state-of-the-art results on the RIA and Lenta datasets of Russian news. BertSumAbs increases ROUGE on average by 2.9 and 2.0 points respectively over previous best score achieved by Phrase-Based Attentional Transformer and CopyNet.

\keywords{Text Summarization \and Headline Generation \and Russian language \and BERT}
\end{abstract}

\section{Introduction}
Text summarization aims to condense vital information from text into a shorter, coherent form that includes main ideas. Two main approaches are distinguished: extractive, which involves organizing words and phrases extracted from text to create a summary, and abstractive, which requires the ability to generate novel phrases not featured in the source text while preserving the meaning and essential information.

Headline generation is considered within the automatic text generation area, so these methods are conventional approaches to that task. Because headlines are usually shorter than summaries, the model has to be good at distinguishing the most salient theme and compressing it in a syntactically correct way. The task is vital for news agencies and especially news aggregators~\cite{ref_news_aggr}. The right solution can be beneficial both for them and for the readers. The headline is the most widely read part of any article, and due to its summarization abilities, it can help decide whether a particular article is worth spending time.

An essential property of the headline generation task is data abundance. It is much easier to collect a dataset in any language containing articles with headlines than articles with summaries because articles usually have a headline by default. We can use the headline generation as a pretraining phase for other problems like text classification or clustering news articles. This two-stage fine-tuning approach is shown to be effective~\cite{ref_bert_ft}. That is why it is essential to investigate the performance of different models on this particular task.

In this paper, we explore the effectiveness of applying the pretrained Trans-former-based models for the task of headline generation. Concretely, we fine-tune mBART and BertSumAbs models and analyze their performance. We obtain results that validate the applicability of these models to the headline generation task.

\section{Related Work}

Previous advances in abstractive text summarization have been made using RNNs with an attention mechanism~\cite{ref_att}.

One more important technique to improve RNN encoder-decoder model is copying mechanism~\cite{ref_copynet} that increases the model ability to copy tokens from the input. LSTM-based CopyNet on byte pair encoded tokens achieved the previous state-of-the-art results on Lenta dataset~\cite{ref_headline_copynet}. In the Pointer-Generator network (PGN) that idea was further developed by introducing coverage mechanism~\cite{ref_pgn} to keep track of what has been summarized.

The emergence of pretrained models based on Transformer architecture~\cite{ref_trans} led to new improvements. Categorization, history and applications of these models are comprehensively described in the survey~\cite{ref_trans_survey}. Applying a Phrase-Based Attentional Transformer (PBATrans) achieved the latest state-of-the-art results on RIA dataset that we are also considering~\cite{ref_lasota}.

\section{Models Description}

Applying Transformer-based models usually follows these steps. Firstly, during unsupervised training, the model learns the universal representation of language. Then it is fine-tuned to a downstream task. Below we briefly describe the pretrained models we focus on in this work.

The first model is mBART~\cite{ref_mbart}. It is a standard Transformer-based model consisting of an encoder and autoregressive decoder. It is trained by reconstructing the document from a noisy version of that document. Document corruption strategies include randomly shuffling the original sentences' order, randomly rotating the document so that it starts from different token and text infilling schemes with different span lengths. mBART is pretrained once for all languages on a subset of 25 languages. It has 12 encoder layers and 12 decoder layers with a total of $\sim$680M parameters, and its vocabulary size is 250K.

The second model we examine is BertSumAbs~\cite{ref_presum}. It utilizes pretrained Bidirectional Encoder Representations from Transformers (BERT)~\cite{ref_bert} as a building block for the encoder. The encoder itself is a 6 stacked layers of BERT. Multi-sentence representations [CLS] tokens are added between sentences, and interval segmentation embeddings are used to distinguish multiple sentences within a document. The decoder is randomly initialized 6-layered Transformer. However, because the encoder is pretrained and the decoder is trained from scratch, the fine-tuning may be unstable. Following~\cite{ref_presum}, we use a fine-tuning schedule that separates the optimizers to handle this mismatch.

As a pretrained BERT for BertSumAbs, we use RuBERT trained on the Russian part of Wikipedia and news data ~\cite{ref_rubert}. A vocabulary of Russian subtokens of size 120K was built from this data as well. BertSumAbs has $\sim$320M parameters. 

Scripts and models' checkpoints for both models are available\footnote{https://github.com/leshanbog/PreSumm}.

\section{Datasets}

The RIA dataset consists of news articles and associated headlines for around 1 \nolinebreak million examples~\cite{ref_sa_ria}. These documents were published on the website ”RIA Novosti (RIA news)” from January 2010 to December 2014. For training purposes, we split the dataset into the train, validation, and test parts in a proportion of 90:5:5.

The Lenta dataset\footnote{https://github.com/yutkin/Lenta.Ru-News-Dataset} contains about 800 thousand news articles with titles that were published from 1999 to 2018. The purpose of model evaluation on this dataset is to measure the model capabilities to generate summaries given articles with another structure, different period, and style of writing.

There are no timestamps in both datasets, so time-based splits are unavailable. It can cause some bias because models tend to perform better on texts with entities that they saw during training. 

\section{Experiments}
\subsection{Evaluation}
\label{eval_sect}
To evaluate the model, we use the ROUGE metric~\cite{ref_rouge}, reporting $F_1$ score for unigram and bigram overlap (ROUGE-1, ROUGE-2), and the longest common subsequence (ROUGE-L). We use the mean of these three metrics as the primary metric (R-mean). To balance this with a precision-based measure, we report BLEU~\cite{ref_bleu}.

We also reporting the proportion of novel n-grams that appear in the summaries but not in the source texts. The higher that number is the fewer copying was made to generate a summary meaning more abstractive model.

As baseline models we use first sentence and PGN. First sentence is a strong baseline for news articles because often the most valuable information is at the start while further sentences provide details and background information. PGN on byte-pair encoded tokens approach was described in the paper \cite{ref_headline_copynet}. The implementation and parameters for PGN are used from here\footnote{https://github.com/IlyaGusev/summarus}.

\subsection{Training dynamics}

\begin{figure}
\centering
\includegraphics[width=0.75\textwidth]{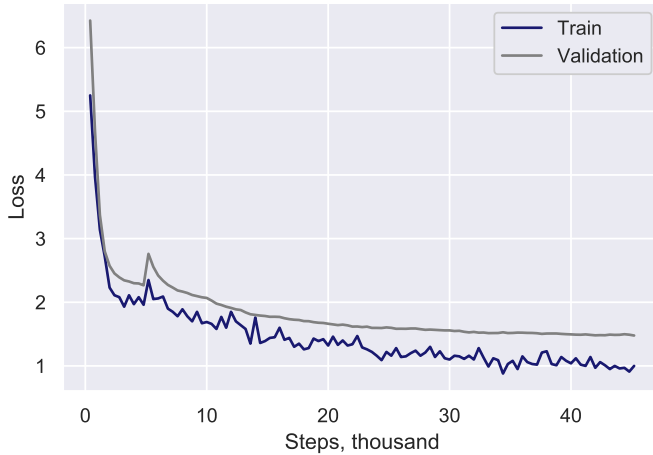}
\caption{BertSumAbs training dynamics on train and validation parts} \label{fig1}
\end{figure}

In Fig.~\ref{fig1} we present the training dynamics of the BertSumAbs model. It takes about 3 days to train for 45K steps (on GeForce GTX 1080) with a batch size equals 8, gradient accumulation every 95 steps. We use a 40K checkpoint as a final model due to its better loss score on the validation dataset.

\begin{table}
\centering
\begin{tabular}{| p{2.5cm} |  p{1.3cm} | p{1.3cm} | p{1.3cm} | p{1.3cm}| p{1.3cm} |}
\hline
Model & R1  & R2 & RL & R-mean & BLEU\\
\hline
\hline
First sentence & 23.8  & 10.5  & 16.6 & 16.9 & 21.8\\
\hline
CopyNet~\cite{ref_headline_copynet} & 41.6  & 24.5  & 38.9  & 35.0 & 53.8\\
\hline
PBATrans~\cite{ref_lasota} & 43.0 & 25.4 & 40.0 & 36.1 & -\\
\hline
PGN & 42.3  & 25.1  & 39.6  & 35.7 & 54.2\\
\hline
mBART & 42.8 & 25.5 & 39.9 & 36.1 & 55.1\\
\hline
BertSumAbs & \bfseries 46.0 & \bfseries 28.0 & \bfseries 43.1 & \bfseries 39.0 & \bfseries 57.6\\
\hline
\end{tabular}
\newline
\caption{RIA dataset evaluation}\label{tab1}
\end{table}
\section{Results}
Table~\ref{tab1} demonstrates the evaluation results on the RIA dataset. There is a significant improvement in all considered metrics for BertSumAbs whereas mBART performance is on the previous state-of-the-art level. Table~\ref{tab2} presents results on the Lenta dataset while models are trained on the RIA dataset. Both models show an improvement compared to previous results for all metrics. On both datasets, BertSumAbs shows better performance than mBART. A possible reason is that BertSumAbs utilizes RuBERT trained specifically on Russian corpus, whereas Russian is one of 25 languages for mBART.

\begin{figure}
\centering
\includegraphics[width=1\textwidth, height=5cm]{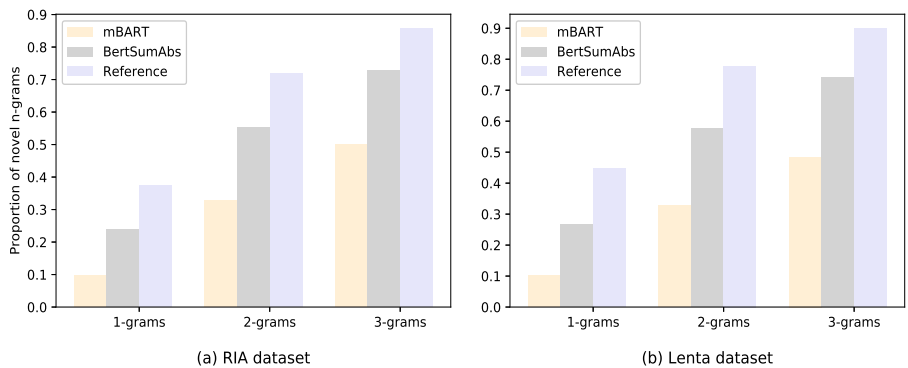}
\caption{Proportion of novel n-grams in headlines} \label{fig2}
\end{figure}

\begin{table}
\centering
\begin{tabular}{| p{2.5cm} |  p{1.3cm} | p{1.3cm} | p{1.3cm} | p{1.3cm}| p{1.3cm} |}
\hline
Model & R1  & R2 & RL & R-mean & BLEU\\
\hline
\hline
First sentence & 24.0  & 10.6  & 18.3 & 17.6 & 24.9\\
\hline
CopyNet~\cite{ref_headline_copynet} & 28.3  & 14.0  & 25.8  & 22.7 & 40.4\\
\hline
PGN & 26.4 & 12.3 & 24.0 & 20.9 & 39.8\\
\hline
mBART & 30.3 & 14.5 & 27.1 & 24.0 & 43.2\\
\hline
BertSumAbs &  \bfseries 31.0  & \bfseries 14.9 &  \bfseries 28.1 & \bfseries 24.7 &  \bfseries 45.1\\
\hline
\end{tabular}
\newline 
\caption{Lenta dataset evaluation using model trained on RIA dataset}\label{tab2}
\end{table}
Fig.~\ref{fig2} presents the proportion of novel n-grams for both models in comparison with true headlines, designated here and below as Reference. Results show that BertSumAbs is more abstractive on both datasets. A manual inspection confirms the fact that mBART is more prone to coping.

To measure the impact of time bias, we scrape about 2.5K articles and associated headlines from the RIA website\footnote{https://ria.ru/}. All articles were published in the 2020 year. The results are shown in Table~\ref{tab3}. As expected, there is a decrease in metrics which is explained by new entities in articles. BertSumAbs suffers a greater decrease than mBART which may be because of more diverse pretraining of the latter.

\begin{table}
\centering
\begin{tabular}{| p{2.5cm} |  p{1.3cm} | p{1.3cm} | p{1.3cm} | p{1.3cm}| p{1.3cm} |}
\hline
Model & R1  & R2 & RL & R-mean & BLEU\\
\hline
\hline
PGN & 39.6 & 20.8 & 35.2 & 31.9 & 52.1\\
\hline
mBART & 41.7 & 22.7 & 37.2 & 33.9 & 53.2\\
\hline
BertSumAbs & 41.9 & 22.5 & 37.3 & 33.9 & 54.2\\
\hline
\end{tabular}
\newline 
\caption{Evaluation on RIA articles from 2020}\label{tab3}
\end{table}

\begin{table}[hbt!]
\centering
\begin{tabular}{ | p{2.3cm} | p{9.5cm} |}
\hline
1. Reference & курс доллара подрос на открытии в среду на 1 коп - до 28,04 руб\\
 & en: \textit{the dollar rate increased at the opening on wednesday by 1 kopeck - up to 28,04 rubles}\\
\hline
1. mBART & рубль укрепился на открытии на открытии на открытии на одну копейку\\
 & en: \textit{ruble strengthened at the opening at the opening at the opening by 1 kopeck}\\
\hline
1. BertSumAbs & курс доллара подрос на открытии в среду на 1 коп - до 28,04 руб\\
& en: \textit{the dollar rate increased at the opening on wednesday by 1 kopeck - up to 28,04 rubles}\\
\hline
\hline
2. Reference & 7 дней в гаване: неделя в кубе\\
& en: \textit{7 days in havana: a week in cuba}\\
\hline
2. mBART & "экстранхерос" или "виахерос"\\
& en: \textit{"extranheros" or "viheros"}\\
\hline
2. BertSumAbs & кубинская мама, или как таксисты застряли на таможне\\
& en: \textit{сuban mom, or how taxi drivers got stuck at customs}\\
\hline
\hline
3. Reference & женщина без рук стирает ногами, а зубами развешивает белье\\
& en: \textit{a woman without hands washes her feet and hangs clothes with her teeth}\\
\hline
3. mBART & безрукой женщине, живущей в калужской области, говорят: "имела бы ты руки"\\
& en: \textit{to armless woman living in the kaluga region say: "would you have hands"}\\
\hline
3. BertSumAbs & безрукая женщина в калужской области родила безрукую женщину\\
& en: \textit{armless woman in the kaluga region gave birth to an armless woman}\\
\hline
\hline
4. Reference & китаец четыре года жил с ножом в голове, не зная об этом\\
& en: \textit{for four years a chinese man lived with a knife in his head, not knowing about it}\\
\hline
4. mBART & врачи обнаружили лезвие ножа в голове своего пациента\\
& en: \textit{doctors found a knife blade in their patient’s head}\\
\hline
4. BertSumAbs & врачи обнаружили в голове жителя китая десятисантиметровое ружьe\\
& en: \textit{doctors found in the head of a resident of china a ten-centimeter gun}\\
\hline
\end{tabular}
\newline
\caption{Bad examples}\label{tab4}
\end{table}

\subsection{Human Evaluation}

To understand model performance beyond automatic metrics, we conducted the human evaluation of the results. We randomly sample 1110 articles and show them both with true and predicted headlines to human annotators. We use Yandex.Toloka\footnote{https://toloka.yandex.com/} to manage the process of distributing samples and recruiting annotators. The task was to decide which headline is better: generated by BertSumAbs or by human. There was also a draw option if the annotator could not decide which headline is better. There was 9 annotators for each example and they did not know about the origin of the headlines. As a result, we get draw in 8\% and BertSumAbs win in 49\% of samples. Analysing the aggregate statistics, we found that BertSumAbs headlines were chosen by 5 or more annotators in 32\% whereas human generated in 28\%. 
\subsection{Error Analysis}
Although, in general, the model's output is fluent and grammatically correct, some common mistakes are intrinsic for both models. In Table~\ref{tab4}, we provide several such examples. In the first example, there is a repeated phrase in mBART output, while BertSumAbs is very accurate. The second example is so hard for both models that there are imaginary words in mBART prediction. It is a common problem for models operating on a subword level. The third example confirms that some articles are hard for both models. One more type of mistake is factual errors, as in the fourth example, where BertSumAbs reported about the wrong subject. This type of error is the worst because it is the most difficult to detect.

\section{Conclusion and Future Work}

In this paper, we showcase the effectiveness of fine-tuning the pretrained Trans-former-based models for the task of abstractive headline generation and achieve new state-of-the-art results on Russian datasets. We showed that BertSumAbs that utilize language-specific encoder has better results than mBART. Moreover, human evaluation confirms that BertSumAbs is capable of generating headlines indistinguishable from the human-created ones.

In future work, we should use the headline generation task as a first step in the two-stage fine-tuning strategy to transfer knowledge to other tasks, such as news clustering and classification.

%
%
%

\begin{thebibliography}{8}


\bibitem{ref_news_aggr}
Murao, K., Kobayashi, K., Kobayashi, H., Yatsuka, T., Masuyama, T., Higurashi, T. and Tabuchi, Y., 2019, June. A Case Study on Neural Headline Generation for Editing Support. In Proceedings of the 2019 Conference of the North American Chapter of the Association for Computational Linguistics: Human Language Technologies, Volume 2 (Industry Papers) (pp. 73-82).

\bibitem{ref_bert_ft}
Sun, C., Qiu, X., Xu, Y. and Huang, X., 2019, October. How to fine-tune BERT for text classification?. In China National Conference on Chinese Computational Linguistics (pp. 194-206). Springer, Cham.


\bibitem{ref_att}
Rush, A.M., Chopra, S., Weston, J.: A neural attention model for abstractive
sentence summarization. In: Empirical Methods in Natural Language Processing.
pp. 379–389 (2015)

\bibitem{ref_trans}
Ashish Vaswani, Noam Shazeer, Niki Parmar, Jakob Uszkoreit, Llion Jones, Aidan N. Gomez, Lukasz Kaiser, and Illia Polosukhin. Attention is all you need. In NeurIPS, 2017

\bibitem{ref_trans_survey}
Qiu, X., Sun, T., Xu, Y., Shao, Y., Dai, N. and Huang, X., 2020. Pre-trained models for natural language processing: A survey. arXiv preprint arXiv:2003.08271.

\bibitem{ref_mbart}
Liu, Y., Gu, J., Goyal, N., Li, X., Edunov, S., Ghazvininejad, M., Lewis, M. and Zettlemoyer, L., 2020. Multilingual denoising pre-training for neural machine translation. arXiv preprint arXiv:2001.08210.


\bibitem{ref_presum}
Liu, Yang, and Mirella Lapata. “Text Summarization with Pretrained Encoders.” Proceedings of the 2019 Conference on Empirical Methods in Natural Language Processing and the 9th International Joint Conference on Natural Language Processing (EMNLP-IJCNLP) (2019)


\bibitem{ref_bert}
Devlin, J., Chang, M.W., Lee, K. and Toutanova, K., 2018. Bert: Pre-training of deep bidirectional transformers for language understanding. arXiv preprint arXiv:1810.04805.

\bibitem{ref_rubert}
Kuratov, Y. and Arkhipov, M., 2019. Adaptation of deep bidirectional multilingual transformers for russian language. arXiv preprint arXiv:1905.07213.


\bibitem{ref_lasota}
Sokolov A, 2019. Phrase-Based Attentional Transformer For Headline Generation. Computational Linguistics and Intellectual Technologies: Proceedings of the International Conference “Dialogue 2019”



\bibitem{ref_sa_ria}
Gavrilov, D., Kalaidin, P. and Malykh, V., 2019, April. Self-attentive model for headline generation. In European Conference on Information Retrieval (pp. 87-93). Springer, Cham.


\bibitem{ref_headline_copynet}
Gusev, I.O., 2019. Importance of copying mechanism for news headline generation, in: Komp’juternaja Lingvistika i Intellektual’nye Tehnologii. ABBYY PRODUCTION LLC, pp. 229–236.


\bibitem{ref_rouge}
Lin, C.Y. and Och, F.J., 2004, June. Looking for a few good metrics: ROUGE and its evaluation. In Ntcir Workshop.

\bibitem{ref_bleu}
Papineni, K., Roukos, S., Ward, T., Zhu, W. J.: BLEU: a method for automatic evaluation of machine translation, 40th Annual meeting of the Association for Computational Linguistics, pp. 311–318 (2002).



\bibitem{ref_copynet}
Gu, J., Lu, Z., Li, H. and Li, V.O., 2016. Incorporating copying mechanism in sequence-to-sequence learning. arXiv preprint arXiv:1603.06393.

\bibitem{ref_pgn}
See, A., Liu, P., Manning, C.: Get To The Point: Summarization with Pointer-Generator Networks. In: Proceedings of the 55th Annual Meeting of the Associationfor Computational Linguistics, vol.1, pp. 1073–1083, Association for ComputationalLinguistics, Vancouver (2017)



\end{thebibliography}
%

\end{document}